 \documentclass[pmlr,twocolumn,10pt]{jmlr} 

\usepackage{booktabs}
\usepackage[most]{tcolorbox}
\usepackage{graphicx}
\usepackage{multirow}

\usepackage{siunitx}

\usepackage[switch]{lineno}

\theorembodyfont{\upshape}
\theoremheaderfont{\scshape}
\theorempostheader{:}
\theoremsep{\newline}

\jmlrvolume{259}
\jmlryear{2024}
\jmlrsubmitted{LEAVE UNSET}
\jmlrpublished{LEAVE UNSET}
\jmlrworkshop{Machine Learning for Health (ML4H) 2024} 

\title[Path-RAG: Knowledge-Guided Key Region Retrieval for Open-ended Pathology VQA]{Path-RAG: Knowledge-Guided Key Region Retrieval for Open-ended Pathology Visual Question Answering}

\author{%
\Name{Awais Naeem} \Email{awais.naeem@utexas.edu}\\
\addr University of Texas at Austin, USA
\AND
\Name{Tianhao Li}\Email{tianhao@utexas.edu}\\
\addr University of Texas at Austin, USA
\AND
\Name{Huang-Ru Liao} \Email{mimiliao2000@utexas.edu}\\
\addr University of Texas at Austin, USA
\AND
\Name{Jiawei Xu} \Email{jiaweixu@utexas.edu}\\
\addr University of Texas at Austin, USA
\AND
\Name{Aby Mammen Mathew} \Email{abymmathew@utexas.edu}\\
\addr University of Texas at Austin, USA
\AND
\Name{Zehao Zhu} \Email{zehaozhu@utexas.edu}\\
\addr University of Texas at Austin, USA
\AND
\Name{Zhen Tan} \Email{ztan36@asu.edu}\\
\addr Arizona State University, USA
\AND
\Name{Ajay Kumar Jaiswal} \Email{ajayjaiswal@utexas.edu}\\
\addr University of Texas at Austin, USA
\AND
\Name{Raffi A. Salibian} \Email{rsalibian@aol.com}\\
\addr University of California, Los Angeles, USA
\AND
\Name{Ziniu Hu} \Email{bull@cs.ucla.edu}\\
\addr University of California, Los Angeles, USA
\AND
\Name{Tianlong Chen} \Email{tianlong@mit.edu}\\
\addr Massachusetts Institute of Technology, USA
\AND
\Name{Ying Ding} \Email{ying.ding@ischool.utexas.edu }\\
\addr University of Texas at Austin, USA
}

\begin{document}

\maketitle              
\begin{abstract}
Accurate diagnosis and prognosis assisted by pathology images are essential for cancer treatment selection and planning. Despite the recent trend of adopting deep-learning approaches for analyzing complex pathology images, they fall short as they often overlook the domain-expert understanding of tissue structure and cell composition. In this work, we focus on a challenging Open-ended Pathology VQA (PathVQA-Open) task and propose a novel framework named Path-RAG, which leverages HistoCartography to retrieve relevant domain knowledge from pathology images and significantly improves performance on PathVQA-Open. Admitting the complexity of pathology image analysis, Path-RAG adopts a human-centered AI approach by retrieving domain knowledge using HistoCartography to select the relevant patches from pathology images. Our experiments suggest that domain guidance can significantly boost the accuracy of LLaVA-Med from 38\% to 47\%, with a notable gain of 28\% for H\&E-stained pathology images in the PathVQA-Open dataset. For longer-form question and answer pairs, our model consistently achieves significant improvements of 32.5\% in ARCH-Open PubMed and 30.6\% in ARCH-Open Books on H\&E images. All our relevant codes and datasets will be open-sourced.


\end{abstract}

\paragraph*{Data and Code Availability}
We use PathVQA~\cite{he2020pathvqa} dataset and introduce a custom dataset namely ARCH-Open (https://github.com/embedded-robotics/path-rag/tree/main/ARCH-OPEN), both of which are publicly available to other researchers for use and reproducibility. The code for PathRAG implementation is available here: https://github.com/embedded-robotics/path-rag.

\paragraph*{Institutional Review Board (IRB)}
This research does not require IRB approval as it does not involve human subjects or data that would necessitate IRB oversight.

\section{Introduction}
Pathology image analysis plays an essential role in the diagnosis, treatment, and study of cancer~\cite{Wang2019-vm}. For example, reading hematoxylin and eosin (H\&E)-stained slides is a clinical standard of care for the diagnosis and staging of many cancers~\cite{Luo2017-jq}. The advent of whole-slide image (WSI) scanning, capable of capturing intricate histological details at high resolutions, has enabled remarkable advancements in pathology image analysis through deep learning algorithms to detect cancer regions~\cite{Kather2016-fn}, classify tissue subtypes~\cite{Javed2020-cg}, and identify diagnostically relevant structures and regions (e.g., nuclei~\cite{Gamper2019-fr}, glands~\cite{Graham2019-ux}, vessels and nerves~\cite{Fraz2020-dx}).

With recent advance of foundation model, a natural thinking is to utilize Visual Language Models (VLMs), like GPT4-V and LLAVA, for helping pathology analysis. However, studies show that on Open-ended Pathology visual question answering (VQA), recent state-of-the-art VLM LLAVA, even after fine-tuning, only achieves a recall of 38\%~\cite{li2024llava}. 

The reason is because these VLMs are hard to identify fine-grained visual objects and textual entities, especially the intricate domain knowledge in pathology images. Therefore, in this paper, we resort to domain knowledge in pathology literature to help narrow down and identify key regions, and then use VLM as soft knowledge base to retrieve most important information from the image to be diagnosed. Specifically, we propose Path-RAG, refer to \underline{R}etrieval \underline{A}ugmentation \underline{G}eneration for \underline{Path}ology VQA.

\begin{figure*}[t]
    \centering
    \vspace{-1.5em}
    \includegraphics[width=1\linewidth]{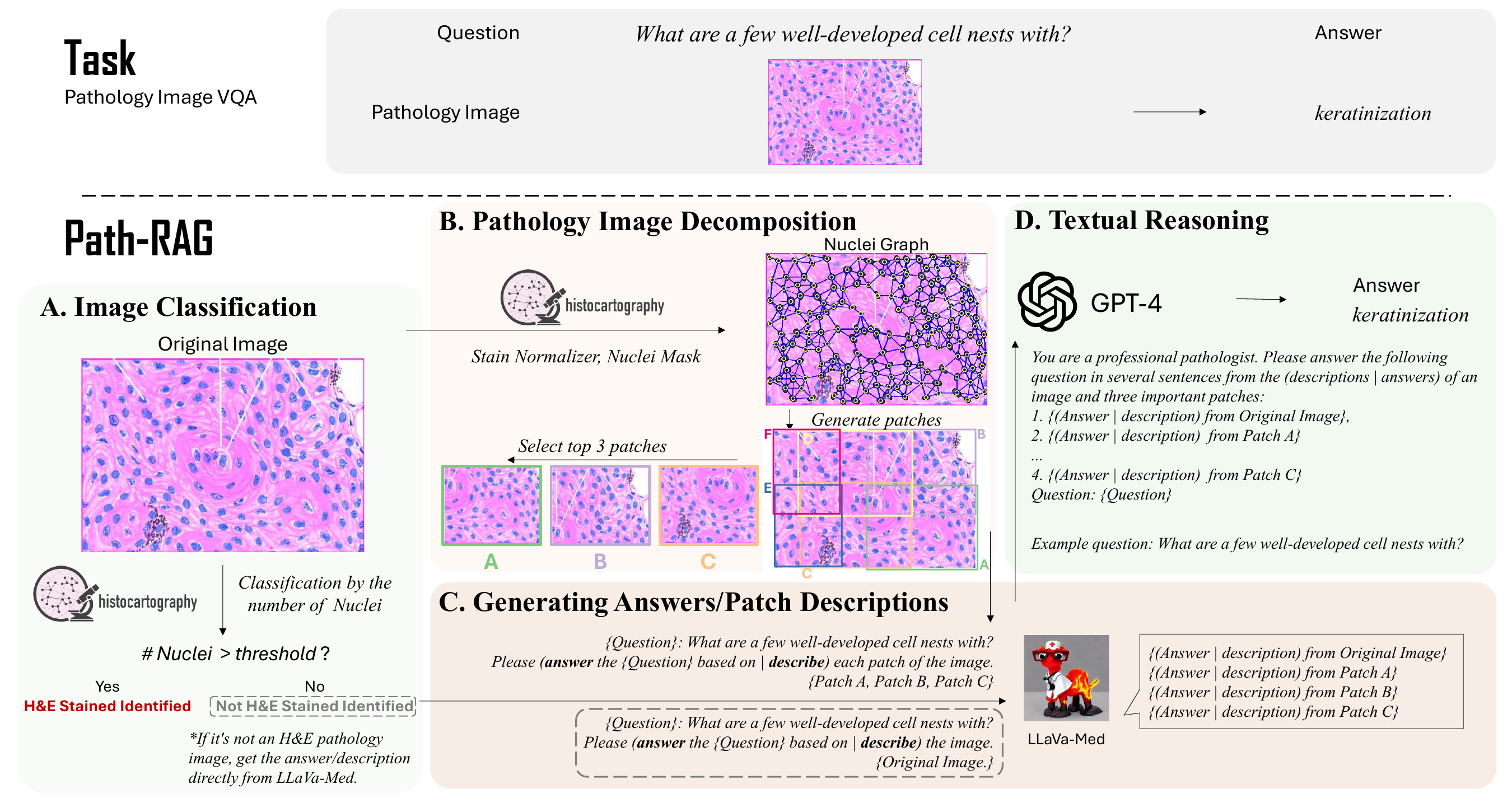}
    \caption{The overview of our Path-RAG framework.}
    \vspace{-1.5em}
    \label{fig:Path-RAG}
\end{figure*}

The key question is which domain knowledge we'd like to retrieve.
Since pathology images are complex and tightly connected with different tissue structures, normal deep learning approaches, that typically involve patch generation, patch-level feature extraction, and feature aggregation to generate image-level embeddings for pathology tasks, fall short as they often overlook crucial tissue structure and cell composition~\cite{pmlr-v156-jaume21a}. The critical step for diagnosing pathology images is the pathologist’s ability to identify tissue structure and examine the cellular composition, organization, and architecture of tissues to characterize features of normal structures and abnormal cells in histologic sections~\cite{Bauer2020-ka}.

For analyzing this, cell graphs of cancer have been generated based on pathology images showing the distinct features compared with healthy cells at the cellular level and at the tissue level~\cite{Gunduz2004-ay}. Cell graphs are one kind of entity graph where nodes and edges of the graphs denote tissue entities and their interactions. The entities can be nuclei, tissue regions, and glands. Entity graphs can simultaneously capture local entity environment and global tissue composition, illustrating an ideal way to capture domain-specific knowledge in pathology images. Entity graphs can improve the efficiency in diagnosis considering that only diagnostically relevant entities have been analyzed, instead of analyzing the entire pathology images~\cite{Shaban2020-ny}. Generating entity graphs requires a complex workflow including stain normalization, tissue detection, entity detection, entity encoding, and constructing graph topology. HistoCartography is an open-source Python library that unifies a set of histology image manipulation tools to build entity graphs~\cite{Graham2019-ux,Ahmedt-Aristizabal2022-sa}, including stain normalization, tissue detection, nuclei detection, tissue component detection, feature extraction, and cell graph builders.

In Path-RAG, we utilize the existing HistoCartography tool to help decompose given image into several key region patches, and then use LLAVA-Med to summarize key information as well as candidate answer for each patch. The retrieved knowledge are fed into the final GPT-4 model for reasoning to provide the answer. In addition to using LLAVA-MED fine-tuned on PathVQA~\cite{he2020pathvqa}, we also fine-tune LLAVA-MED on a custom dataset consisting of detailed open-ended question-answer pairs and pathology images which aids us in evaluating the efficacy of our knowledge-guided approach on detailed answers unlike PathVQA where majority of the open-ended answers contain just 2-3 words. Our primary contributions can be summarized as:

\vspace{-0.3em}
\begin{itemize}
    \item[$\bullet$] \textbf{Significant Performance Improvement with up to 28\% increase in the recall.}
    Path-RAG is a human-centered AI approach by retrieving domain knowledge using HistoCartography to select the most relevant patches for a given pathology image, which can significantly boost the recall of LlaVa-Med for PathVQA Open-ended questions from 38\% to 47\%, especially with a significant improvement of 28\% for H\&E-stained pathology images. For longer-form question and answer pairs, our model consistently achieves a significant improvement of 32.5\% in ARCH-Open PubMed and 30.6\% in ARCH-Open Books on H\&E images.
    \vspace{0.1em}
    \item[$\bullet$] \textbf{Consistent performance gain with and without GPT-4.} Path-RAG has demonstrated consistent performance benefits over LLaVa-Med for open-ended QA with and without GPT-4 component indicating the significance of domain guidance in pathology image analysis.
    \vspace{0.1em}
    \item[$\bullet$] \textbf{Performance improvement for non-finetuned LLaVa-Med.}
    Path-RAG can improve the recall by 12\% when using LLaVa-Med not finetuned on PathVQA to generate descriptions as a prompt for GPT-4 to get the answer on H\&E-stained slides.
    \vspace{0.1em}
    \item[$\bullet$] \textbf{Dataset having detailed Open-Ended Pathology Visual Question-Answer Pairs.}
    ARCH-Open dataset contains extensive open-ended question-answer pairs about pathology images which can be used for VQA systems and training/fine-tuning multi-modal LLMs for pathology related tasks.
\end{itemize}


\section{Methodology}

In our approach to handling pathology VQA, we are employing a careful ensemble of tools built specifically to process pathology images with large language models (LLMs) pre-trained on the pathology data. Our approach obviates any need to further fine-tune LLMs which incur high computational and memory overhead. We are \textit{the first} to illustrate that using an off-the-shelf domain-aware tool can significantly benefit the intricate task of pathology open-ended VQA rather than investing heavily in compute resources, which happens to be the norm when dealing with multi-modal data in the pathology domain. Furthermore, we convert the multi-modal problem (including image and text) into a solely text-based problem and then use the textual reasoning capabilities of GPT-4 as the last step. This technique gives us the leverage to use multiple LLMs; one fine-tuned to handle pathology VQA in a multi-modal setting~\cite{li2024llava} and another one specialized for textual reasoning~\cite{openai2024gpt4}.

Pathology images can be categorized and represented in discrete units such as tissues, nuclei, cells, glands, etc. To answer any question related to pathology, it is considered essential to understand the image regions which are heavily concentrated in any discrete unit under consideration. In fact, this approach is used by radiologists when they are reasoning about the question asked about any diagnostic image~\cite{Bauer2020-ka}. To follow a similar practice of analyzing pathology images, we are utilizing HistoCartography~\cite{pmlr-v156-jaume21a} which is a toolkit containing pre-trained models for graph analytics in digital pathology. We are using this tool for two purposes: (1) Differentiate between pathology and non-pathology images (2) Extract image patches which are concentrated in nuclei structures.

LLaVA-Med ~\cite{li2024llava} is a multi-modal language model fine-tuned on biomedical data and exhibits conversational capabilities to answer open-ended questions about a biomedical image. Since it can follow open-ended instructions to give relevant details about any biomedical image, we employ LLaVA-Med to generate descriptions/answers about the pathology image and/or its relevant patches. Finally, we use GPT-4~\cite{openai2024gpt4} to reason about the question using the textual descriptions generated via LLaVA-Med.

Our method uses a sequential approach to use these components as shown in Figure 1 and  subsequently detailed in the following sections.

\subsection{H\&E stained Pathology Image Identification}
As our approach relies on generating the patches of the pathology image depending on the presence of nuclei in the image, firstly we need to determine whether an image is an H\&E stained pathology image or not. Using state-of-the-art approaches, a binary image classifier can be used to differentiate pathology images from non-pathology images. However, to avoid the use of an extra classifier, we have used HistoCartography~\cite{pmlr-v156-jaume21a} tool to determine this classification dynamically. We use a nuclei extractor API in HistoCartography to detect the number of nuclei in the image. For any non-pathology image, the extractor will result in zero or miniature number of nuclei-like components in the image. For our use case, we have used a hard threshold value of minimum 5 nuclei to differentiate between pathology and non-pathology image. Identifying any image as non-pathology gives us leverage to not use HistoCartography to divide this image into patches and directly advance towards LLaVA-Med to get the description/answer about the whole image. For the pathology images, however, the next step is to extract the relevant patches. This approach aligns with our focus on cancer-related pathology, since H\&E stained images are gold standard for tumor diagnosis.

\subsection{Pathology Image Decomposition}
To extract the relevant patches of pathology images, we use different tools available in HistoCartography~\cite{pmlr-v156-jaume21a}. Firstly, we use stain normalizer to overcome stain variability in the input image. Secondly, we detect nuclei present in the processed image using a nuclei extractor module to come up with the pixel locations of nuclei and their centers. Third, a deep feature extractor based on ResNet34~\cite{he2016deep} is used to extract the features of each nucleus by taking input from all the neighborhood image patches around each nucleus~\cite{pmlr-v156-jaume21a}. The encoded nuclei features are then fed into the KNNGraphBuilder which builds a graph based on DGL~\cite{wang2019dgl} such that each nucleus is connected to 5 nearest neighbors with a maximum distance threshold of 50 pixels as per the configuration defined in HistoCartography~\cite{pmlr-v156-jaume21a} tool. In the resulting nuclei graph, each nuclei center is represented by node whereas the edges connect the nearest neighbor nuclei. The procedure is illustrated in part B of Figure~\ref{fig:Path-RAG}.

Once the graph is obtained, a custom approach is taken to extract only those patches of the image which have maximum number of nuclei. In this approach, the pathology image is divided into a total of 9 patches having 20\% overlap between them irrespective of the total dimensions of the image. It results in varying patch size for different images because of overlapping between successive patches up/down and avoids miniature/big patches having fixed size. Finally, all the patches are ranked as per the number of nuclei centers contained in each patch and different numbers of top-ranked patches are considered for experiments and evaluation. To investigate if our domain-aware patch selection strategy is effective, we additionally perform ablation studies with random patches.


\subsection{Patch Captioning and Candidate Answer Extraction}
To generate the textual descriptions and candidate of each patch (we abbreviate as \textit{answer/description} later), we resort to LLaVA-Med~\cite{li2024llava} which can generate an open-ended description of biomedical images. Once the key image regions are extracted as per the domain knowledge~\cite{pmlr-v156-jaume21a}, 
For the pathology images, we provide each patch to LLaVA-Med to \textit{generate the open-ended descriptions}, and then \textit{generate a candidate answer} independently. This results into multiple \textit{answer/description} pairs of each key region along with the original image, which will be later provided to GPT-4 for the final-stage reasoning. 
For the non-pathology images, we only use the input image to generate a single \textit{answer/description} of the complete image. 
The procedure is illustrated in part C of Figure~\ref{fig:Path-RAG}.

\subsection{Reasoning over Per-Patch Extracted Knowledge}
As the last step, we are using GPT-4~\cite{openai2024gpt4} to reason about the final answer given question and all \textit{answer/description} of the patches generated in the last step. For the non-pathology images, we pass in a question along with the complete description of the input image to GPT-4 in order to get a final response. For the pathology images, we have the answers/descriptions of each patch along with the original image, so we consider two different approaches for experimentation/evaluation:
\textit{Path-RAG (answer)} and \textit{Path-RAG (description)}. We pass in candidate answers and descriptions returned by LLaVA-Med, respectively, along with the question to GPT-4 and ask it to reason about the final answer. This response is the final answer which is returned as the final output of the system given a question and a pathology image as an input. The prompts given to GPT-4 for textual reasoning are added in supplmentary material.
We also set a ablation baseline (\textit{Path-RAG w/o GPT-4 (answer)}), which only consider our final answer as the concatenation of different answers returned by LLaVA-Med for each image/patch, without using GPT-4 for reasoning.

\section{Experimental Setup}
\begin{figure}[t]
    \centering
    \includegraphics[width=1\linewidth]{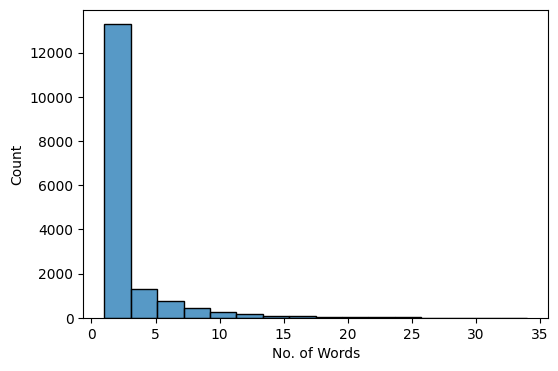}
    \caption{Word Count distribution of open-ended answers in PathVQA}
    \label{fig:PathVQA-Open-Answers}
\end{figure}
\begin{figure}[t]
    \centering
    \includegraphics[width=1\linewidth]{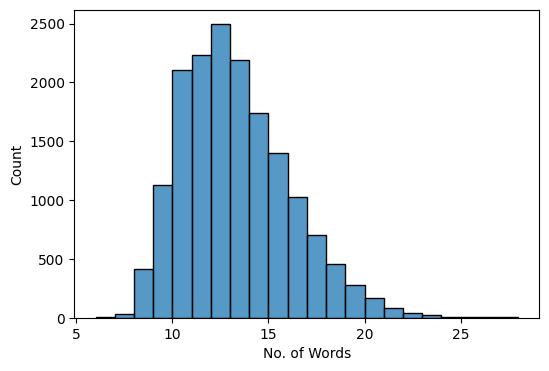}
    \caption{Word Count distribution in ARCH-Open Questions - PubMed}
    \label{fig:ARCH-Open-Questions-PubMed}
\end{figure}
\begin{figure}[t]
    \centering
    \includegraphics[width=1\linewidth]{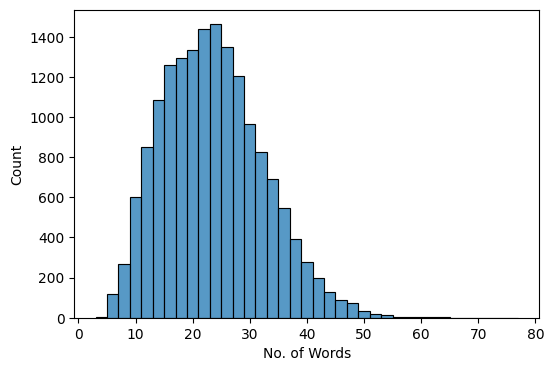}
    \caption{Word Count distribution in ARCH-Open Answers - PubMed}
    \label{fig:ARCH-Open-Answers-PubMed}
\end{figure}
\begin{figure}[t]
    \centering
    \includegraphics[width=1\linewidth]{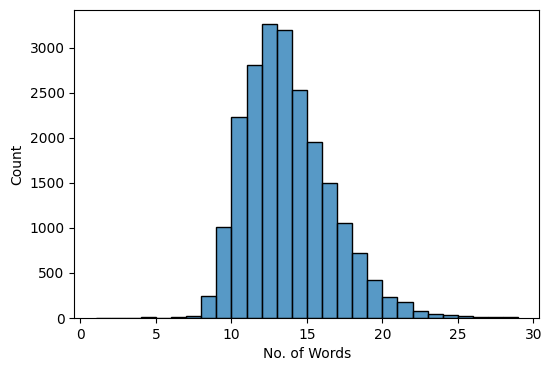}
    \caption{Word Count distribution in ARCH-Open Questions - Books}
    \label{fig:ARCH-Open-Questions-Books}
\end{figure}
\begin{figure}[t]
    \centering
    \includegraphics[width=1\linewidth]{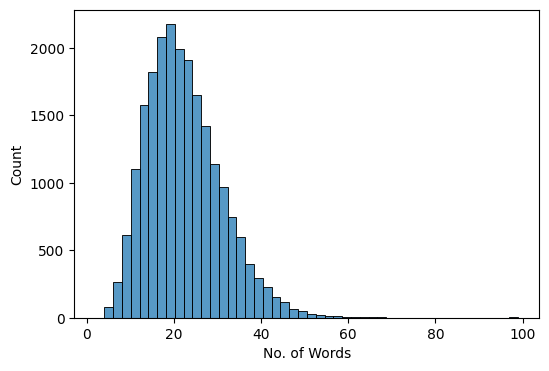}
    \caption{Word Count distribution in ARCH-Open Answers - Books}
    \label{fig:ARCH-Open-Answers-Books}
\end{figure}

\subsection{PathVQA Dataset.}

PathVQA~\cite{he2020pathvqa} contains 32,799 question-answer pairs and 4,998 images from pathology textbooks. Each question has been manually checked to ensure correctness. For each image, several questions may be asked regarding multiple aspects such as location, shape, color, appearance, etc. All of the questions can be categorized into two types by whether the answer is yes or no (close-ended questions) and open-ended questions. We only consider open-ended questions in this work because SoTA methods~\cite{li2024llava, saygin2023quilt} excel at yes/no questions, while performance on open-ended VQA remains low. We use the official training, validation, and test splits in the dataset. Since we do not perform fine-tuning in our work, we only evaluate open-ended questions in the test set to establish consistency with other baselines. There are 849 images and 3,370 questions in the test set, in which 308 images with 1,127 questions are H\&E-stained pathology images and 541 images with 2,243 questions are not H\&E-stained pathology images.

\subsection{ARCH-Open Dataset.} 
\label{sec:ARCH}

To address the lack of datasets accommodating the open-ended nature of pathology VQA, we constructed an open-ended VQA dataset namely ARCH-Open using ARCH~\cite{gamper2020multiple}. ARCH is a multiple instance captioning dataset for pathology images in which caption-image pairs are provided from relevant PubMed articles and books. There are a total of 4,270 captions and 4,305 images from PubMed articles, whereas a total of 3,309 image-caption pairs are extracted from the pathology books.

This dataset addressed the shortcomings of PathVQA~\cite{he2020pathvqa} and was subsequently used to assess the effectiveness of our PathRAG approach along with PathVQA. While PathVQA is the gold standard for open-ended VQA about pathology, it has significant flaws: answers are typically 2-4 words long (Figure: \ref{fig:PathVQA-Open-Answers}), and some questions can be answered without any image. We ensured each ARCH-Open question requires an image for a valid answer and the answer lengths align with the nature of open-ended (long-form) VQA task.

To construct ARCH-Open, we employed instruction tuning by providing image captions to GPT-4~\cite{openai2024gpt4} and then asked it to generate five open-ended question-answer pairs for each image-caption pair considering that caption text belongs to a pathology image (Appendix C). As a result, we got 21,350 VQA pairs from image-caption pairs extracted from PubMed articles (termed as ARCH-Open PubMed), and a total of 16,545 VQA pairs relating to image-caption pairs extracted from pathology books (termed as ARCH-Open Books). In ARCH-Open PubMed, the questions and answers have the average word length of 12.81 and 23.12, respectively (Figure: \ref{fig:ARCH-Open-Questions-PubMed}, \ref{fig:ARCH-Open-Answers-PubMed}), whereas the average word length for questions and answers in ARCH-Open Books is 13.23 and 22.61, respectively (Figure: \ref{fig:ARCH-Open-Questions-Books}, \ref{fig:ARCH-Open-Answers-Books}). Both ARCH-Open PubMed and ARCH-Open Books datasets are split into train/test sets having 80/20 percent instances with only the test set being used for evaluation in our experiments. The quality checks for ARCH-Open are part of our ongoing efforts, involving human pathologists for thorough evaluation.

\subsection{Implementation Settings}

We use LLaVa-Med model and LLaVa-Med model fine-tuned on PathVQA training set to get image descriptions from image input and answers from image and question input. Prompts used are given in Appendix A, B. All inference processes were done on a single NVIDIA RTX A6000 GPU. We use GPT-4 (gpt-4-0125-preview)~\cite{openai2024gpt4} from OpenAI API for the experiments that need to generate final answers after doing inference on  LLaVa-Med model. The temperature of GPT-4 is set to be 0. Because LLMs’ responses are stochastic and depend on the temperature, replication of the scores may be slightly different.




\vspace{-0.5em}
\section{Results and Ablation Study}
\vspace{-0.3em}

\begin{table*}[ht]
\caption{Comparison with prior state-of-the-art supervised methods on PathVQA datasets. Please note that we report our method using 3 patches. w/o GPT-4 (answer) refers to the Path-RAG directly concatenating answers without using GPT-4. (description/answer) refers to different textual input passed to GPT-4 for further reasoning.}
\centering
\footnotesize
\begin{tabular*}{\textwidth}{@{\extracolsep{\fill}}lcccc@{}}
\toprule
\multirow{2}{*}{\textbf{Method}} & \multicolumn{3}{c}{\textbf{Recall}}  \\
\cmidrule(lr){3-5} 
& & \textbf{Not H\&E pathology} & \textbf{H\&E pathology} & \textbf{All} \\
\hline
\multicolumn{5}{c}{\textit{Not Fine-tuned on PathVQA}} \\
\hline
Quilt-LLaVA \cite{saygin2023quilt} & & - & - & 15.3  \\
LLaVA-Med \cite{li2024llava} & & 11.3 & 11.6 & 11.4  \\
\textbf{Path-RAG w/o GPT-4 (answer)}& & 11.3 & 19.2 &  13.9 \\
\textbf{Path-RAG (description)} & & 20.3 & \textbf{28.5} &  \textbf{23.0}   \\
\textbf{Path-RAG (answer)} & & 11.3 & 25.9 &  16.2   \\

\hline
\multicolumn{5}{c}{\textit{Fine-tuned on PathVQA}} \\
\hline
LLaVA-Med \cite{li2024llava} & & 39.0  & 36.4 & 38.1 \\
\textbf{Path-RAG w/o GPT-4 (answer)} & & 39.0 & 51.2 &  43.1   \\
\textbf{Path-RAG (description)} & & 28.7 & 37.0 &  31.5   \\
\textbf{Path-RAG (answer)} & & 39.0 & \textbf{64.1} &  \textbf{47.4}   \\
\bottomrule
\end{tabular*}
\label{tab:sota}
\vspace{-1em}
\end{table*}

\begin{table*}[ht]
\caption{Comparison with prior state-of-the-art supervised methods on ARCH-Open datasets. Please note that we report our method using 3 patches. w/o GPT-4 (answer) refers to the Path-RAG directly concatenating answers without using GPT-4. (description/answer) refers to different textual input passed to GPT-4 for further reasoning.}
\centering
\footnotesize
\begin{tabular*}{\textwidth}{@{\extracolsep{\fill}}lcccc@{}}
\toprule
\multirow{2}{*}{\textbf{Method}} & \multicolumn{3}{c}{\textbf{Recall}}  \\
\cmidrule(lr){3-5} 
& & \textbf{Not H\&E pathology} & \textbf{H\&E pathology} & \textbf{All} \\
\hline
\multicolumn{5}{c}{\textit{Eval on ARCH-Open PubMed}} \\
\hline
LLaVA-Med \cite{li2024llava} & & 53.5 & 52.2 & 52.9  \\
\textbf{Path-RAG w/o GPT-4 (answer)}& & 53.5 & 77.6 & 65.5 \\
\textbf{Path-RAG (description)} & & 75.9 & 80.8 &  \textbf{78.4}   \\
\textbf{Path-RAG (answer)} & & 53.5 & \textbf{84.7} &  69.0   \\
\hline
\multicolumn{5}{c}{\textit{Eval on ARCH-Open Books}} \\
\hline
LLaVA-Med \cite{li2024llava} & & 52.7 & 54.0 & 53.7  \\
\textbf{Path-RAG w/o GPT-4 (answer)}& & 52.7 & 79.5 & 72.3 \\
\textbf{Path-RAG (description)} & & 76.0 & 76.6 &  \textbf{76.4}   \\
\textbf{Path-RAG (answer)} & & 52.7 & \textbf{84.6} &  76.1   \\
\bottomrule
\end{tabular*}
\label{tab:arch_results}
\vspace{-1em}
\end{table*}

\begin{table*}[htbp]
    \centering
    \caption{Effectiveness of patch retrieval methods in pathology images. Note that we report our method using a fine-tuned model. The first column indicates the form of textual input provided to GPT-4 for reasoning. * Indicating the paired bootstrapping values of Histo and Random patches, by sampling with replacement, calculating the mean estimate and confidence interval (CI) [lower bound, upper bound] over 10,000 iterations. }

    \label{tab:patches_effect}
    \begin{tabular}{llccc} 
        \toprule
        \multirow{2}{*}{\textbf{Form}} & \multirow{2}{*}{\textbf{Patches}} & \multicolumn{3}{c}{\textbf{Recall (\%)}}  \\
        \cmidrule(lr){3-5} 
        & & Not H\&E pathology & H\&E pathology & All \\
        \midrule
        Description & No patches & 28.7 & 30.0 & 29.1 \\
        Description & 3 Random patches & 28.7 & 33.9 & 30.4 \\
        Description & 3 Histo patches & 28.7 & 37.0 & 31.5 \\
        & & & \small *(+3.72, CI [1.95, 5.57]) \\
        \midrule
        Answer & No patches & 39.0 & 36.4 & 38.1 \\
        Answer & 3 Random patches & 39.0 & 60.4 & 46.2 \\
        Answer & 3 Histo patches & 39.0 & \textbf{64.1} & \textbf{47.4} \\
        & & & \small *(+3.07, CI [1.13, 5.04]) \\
        \bottomrule
    \end{tabular}
\end{table*}

\begin{table}[htbp]
    \centering
    \small
    \caption{Effect of different number of patches. More patches yield improved performance. Note that we report results using fine-tuned model \& GPT-4 textual reasoning.}
    \label{tab:num_patches}
    \begin{tabular}{lccc}
        \toprule
        \textbf{Form} & \textbf{Num Patches} & \textbf{H\&E-stained} & \textbf{All} \\
        \midrule
        Description & 0 & 30.0 & 29.1 \\
        Description & 3 & 37.0 & 31.5\\
        Description & 6 & 38.3 & 31.9\\
        \midrule
        Answer & 0 & 36.4 & 38.1 \\
        Answer & 3 & 64.1 & 47.4 \\
        Answer & 6 &  \textbf{66.9} &  \textbf{48.4} \\
        \bottomrule
    \end{tabular}
\end{table}

\subsection{Comparison with SOTA:}
In Table \ref{tab:sota}, we compare our method with existing generation-based methods on the open set of PathVQA \cite{he2020pathvqa} datasets. Following previous works ~\cite{li2024llava,saygin2023quilt}, we evaluate our performance using recall. To compare with our main baseline LLaVA-Med~\cite{li2024llava}, we use both finetuned and not finetuned models on PathVQA, and explored different forms of textual input to the GPT-4 model. To highlight the effectiveness of our HistoCartography~\cite{pmlr-v156-jaume21a} patches retrieval, we categorized the testing data into H\&E-stained pathology images or non-H\&E-stained pathology images. Our major findings are: 
\textcircled{1} We achieve SoTA on PathVQA open-set dataset with \textbf{47.4\% recall}. 
\textcircled{2} Our method demonstrates a significant improvement of \textbf{27.7\% over LLaVA-Med} with fine-tuned model on H\&E-stained pathology images, highlighting the importance of incorporating HistoCartography for pathology image analysis. 
\textcircled{3} For models not fine-tuned on PathVQA, our method using description \textbf{outperforms LLaVA-Med by 11.6\%}, providing further evidence of effectiveness of our approach.

In Table \ref{tab:arch_results}, we further evaluate our models on the ARCH-Open dataset. As mentioned in Section \ref{sec:ARCH}, ARCH-Open features longer forms of questions and answers, which better aligns with the real-world setting of VQA. It is worth noting that our method performs consistently well. On H\&E images, we achieve a significant improvement of 32.5\% in ARCH-Open PubMed and 30.6\% in ARCH-Open Books with 3 patches from HistoCartography compared to the baseline \cite{li2024llava} without patches from HistoCartography, demonstrating the effectiveness of our HistoCartography patches retrieval.

\subsection{Effectiveness of Retrieving Patches with HistoCartography:}

To demonstrate the effectiveness of our patches retrieval, we conducted a comparative analysis of three methods: employing no patches, using random patches, and our method, which utilizes domain relevant patches retrieved by HistoCartography~\cite{pmlr-v156-jaume21a} as presented in Table \ref{tab:patches_effect}. For textual input as answer, we first compared scenarios where no patches are employed, our method demonstrates a significant recall improvement of 9.3\%, emphasizing the critical importance of extracting the region of interest patches. These patches provide essential pathology features that can enhance the performance. Secondly, our method outperforms the utilization of random patches by 1.2\% in terms of recall improvement, highlighting the necessity of incorporating domain knowledge to extract relevant patches. Notably, the presence of features such as the number of nuclei in a patch emerges as significant regions of interest for pathology multi-modal models. Such results highlight the significant advantages brought about by retrieving domain relevant patches using HistoCartography. 


\begin{figure}
    \centering
    \includegraphics[width=1\linewidth]{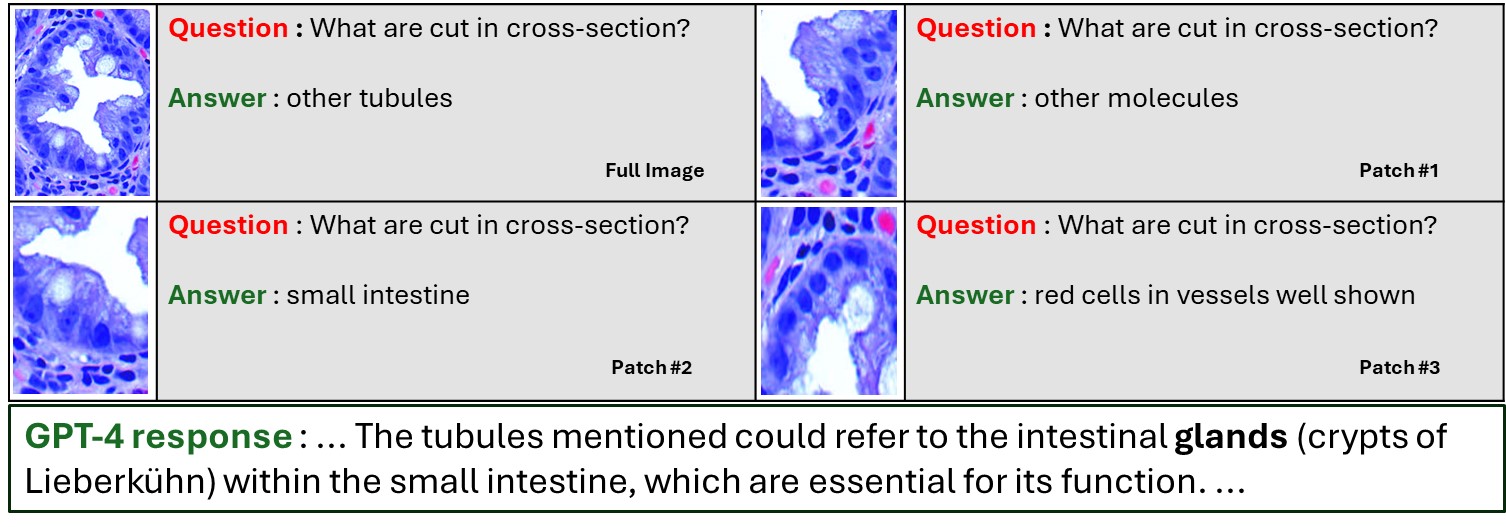}
    \caption{Example overview of Path-RAG with 3 patches. Detailed GPT-4 reasoning to come up with the correct answer (\textbf{bold}) is provided in Appendix A, B.}
    \label{fig:enter-label}
\end{figure}

\subsection{Effect of textual reasoning:}
We ablate the components used for textual reasoning. The results are shown in Table \ref{tab:sota}. Firstly, we investigated the form of textual input to GPT-4. For the fine-tuned model, instructing LLaVA-Med to generate the answer directly and sending it to GPT-4 yields a better recall improvement of $15.9\%$ compared to generating open-ended descriptions and forwarding them to GPT-4. This suggests that leveraging the explicit answer generation capability of fine-tuned LLaVA-Med enhances the effectiveness of GPT-4 in textual reasoning tasks. However, for not fine-tuned model, generating descriptions enhance the recall by $6.8\%$ compared to generating answers directly. This indicates that, in the absence of fine-tuning, providing descriptions as input to GPT-4 may be more effective in facilitating textual reasoning tasks. Secondly, we explored the utilization of GPT-4 for textual reasoning as opposed to directly concatenating the four answers generated by LLaVA-Med. Our results demonstrate that employing GPT-4 for textual reasoning results in a higher recall improvement compared to directly concatenating the answers from different patches for both fine-tuned and not fine-tuned settings. This highlights the superiority of GPT-4's capabilities for textual reasoning tasks using in-context learning, enhancing the performance in open-ended pathology VQA.


\subsection{Effect of different number of patches}
In Table \ref{tab:num_patches}, we evaluate the
impact of employing different numbers of patches. Our findings suggest that incorporating more patches yields improved performance, indicating that employing a greater number of patches is beneficial.

\section{Conclusion}

\vspace{-0.2em}
Open-ended Pathology VQA is a challenging task that requires an understanding of intricate domain knowledge in pathology. Unlike prior works treating this task as simple VQA, our work for the first time argues to inject domain knowledge as guidance using HistoCartography to select the relevant and information-rich patches from pathology images. Our experiments and ablation studies show consistent performance improvement of this retrieval augmentation across multiple design choices. Our future work includes improving PATH-RAG's current patch selection strategies by considering their relevance with open-ended questions. We also aim to explore the performance benefits of Path-RAG for other datasets using vision-language models (GPT4-V~\cite{openai2024gpt4technicalreport}, Qwen2-Vl~\cite{wang2024qwen2vlenhancingvisionlanguagemodels}, LLaVA-Med++~\cite{xie2024medtrinity25mlargescalemultimodaldataset}etc) to capture complex features of the pathology images rather than GPT-4, a text-only model. For a more comprehensive evaluation, we plan to reference recent advancements~\cite{sun2023aligninglargemultimodalmodels} that utilize the GPT-4 model as an oracle to analyze and rate responses across multiple aspects.


\acks{We would like to acknowledge the following funding supports: NIH OT2OD032581, NIH OTA-21-008, NIH 1OT2OD032742-01, NSF 2333703, NSF 2303038.}



%
%
%
\bibliography{references}
\appendix

\section{Path-RAG Prompts}
\begin{tcolorbox}[enhanced jigsaw, breakable, pad at break*=1mm, attach boxed title to top center={yshift=-4mm,yshifttext=-1mm}, 
colback=blue!5!white, colframe=blue!75!black, colbacktitle=gray!80!white, title=Prompts, fonttitle=\bfseries, boxed title style={size=small, colframe=gray!50!black}]
\small
\textcolor[rgb]{0,0,0.9}{Generate Answer from LLaVa-Med:} 
\begin{itemize}
    \item[] \(\langle \text{question} \rangle + \langle \text{image} \rangle\)
\end{itemize}
 
\textcolor[rgb]{0,0,0.9}{Get Response from GPT-4:} You are a professional pathologist. Please generate a comprehensive answer that integrates multiple perspectives for a pathology image and provides a balanced conclusion based on the information provided.

\begin{itemize}
  \item \textcolor[rgb]{0,0,0.9}{Perspective 1}: \textit{Answer from LLaVa-Med with full image input}
  \item \textcolor[rgb]{0,0,0.9}{Perspective 2}: \textit{Answer from LLaVa-Med with Patch 1 input}
  \item \textcolor[rgb]{0,0,0.9}{Perspective 3}: \textit{Answer from LLaVa-Med with Patch 2 input}
  \item \textcolor[rgb]{0,0,0.9}{Perspective 4}: \textit{Answer from LLaVa-Med with Patch 3 input}
  \item \textcolor[rgb]{0,0,0.9}{Question}: \textit{Question}
\end{itemize}

\tcblower
\small
\textcolor[rgb]{0,0,0.9}{Generate Description from LLaVa-Med:} 
\begin{itemize}
    \item[] \(\text{"Describe the following image in detail."} + \langle \text{image} \rangle\)
\end{itemize}

\textcolor[rgb]{0,0,0.9}{Get Response from GPT-4:} You are a professional pathologist. Please answer the question in several sentences based on the description of an image and three important patches.

\begin{itemize}
  \item \textcolor[rgb]{0,0,0.9}{Description of image}: \textit{Description from LLaVa-Med with full image input}
  \item \textcolor[rgb]{0,0,0.9}{Description of patch 1}: \textit{Description from LLaVa-Med with Patch 1 input}
  \item \textcolor[rgb]{0,0,0.9}{Description of patch 2}: \textit{Description from LLaVa-Med with Patch 2 input}
  \item \textcolor[rgb]{0,0,0.9}{Description of patch 3}: \textit{Description from LLaVa-Med with Patch 3 input}
  \item \textcolor[rgb]{0,0,0.9}{Question}: \textit{Question}
\end{itemize}
\end{tcolorbox}

\section{Path-RAG Example}

\begin{tcolorbox}[enhanced jigsaw, breakable, pad at break*=1mm, attach boxed title to top center={yshift=-4mm,yshifttext=-1mm}, 
colback=blue!5!white, colframe=blue!75!black, colbacktitle=gray!80!white, title=Path-RAG GPT-4 Prompt and Response, fonttitle=\bfseries, boxed title style={size=small, colframe=gray!50!black}]
\small
\textcolor[rgb]{0,0,0.9}{Prompt Design:} You are a professional pathologist. Please generate a comprehensive answer that integrates multiple perspectives for a pathology image, providing a balanced conclusion.

\begin{itemize}
  \item \textcolor[rgb]{0,0,0.9}{Perspective 1}: "other tubules"
  \item \textcolor[rgb]{0,0,0.9}{Perspective 2}: "other molecules"
  \item \textcolor[rgb]{0,0,0.9}{Perspective 3}: "small intestine"
  \item \textcolor[rgb]{0,0,0.9}{Perspective 4}: "red cells in vessels well shown"
\end{itemize}

\tcblower
\textcolor[rgb]{0,0,0.9}{Model Response:} 
\scriptsize
In the pathology image under consideration, we see a complex interplay of various elements requiring a detailed analysis from multiple perspectives to reach a comprehensive understanding.

From the first perspective, focusing on "other tubules," we observe structures likely part of the nephron system in the kidney. These tubules, when cut in cross-section, reveal essential architecture for filtration and absorption, suggesting a renal context.

The second perspective, "other molecules," hints at the biochemical aspect. This could involve identifying molecules within the tubules or tissues, indicating metabolic or pathological states.

The third perspective, "small intestine," introduces a gastrointestinal context with villi, essential for nutrient absorption. The structures could represent small intestine tissue if viewed cross-sectionally.

The fourth perspective, "red cells in vessels well shown," emphasizes vascularization, highlighting oxygen and nutrient delivery within the tissue.

In conclusion, the cross-sections likely represent the small intestine with a focus on its internal structure and vascularization. The tubules could represent intestinal glands, with visible molecules indicating enzymatic activity, and clear red cells in vessels underscoring vascular support.
\end{tcolorbox}

\section{Prompt for ARCH-Open Dataset Construction}\label{apd:first}

\begin{tcolorbox}[enhanced jigsaw, breakable, pad at break*=1mm, attach boxed title to top center={yshift=-4mm,yshifttext=-1mm}, 
colback=blue!5!white, colframe=blue!75!black, colbacktitle=gray!80!white, title=ARCH-Open Dataset Prompt, fonttitle=\bfseries, boxed title style={size=small, colframe=gray!50!black}]
\small
\textcolor[rgb]{0,0,0.9}{Prompt Design:} You are provided with a text description (figure caption) of a pathology image. Unfortunately, you don't have access to the original image.
Your job is to generate a total of 5 open-ended question/answer pairs from this figure caption starting with "What" or "Where". Below are the requirements to generate the question/answer pairs:

\begin{itemize}
  \item \textcolor[rgb]{0,0,0.9}{Requirement 1}: Avoid quoting or referring to specific facts, terms, abbreviations, dates, numbers or names, as these may reveal the conversation is based on the text information, rather than image itself
  \item \textcolor[rgb]{0,0,0.9}{Requirement 2}: Focus on the visual aspects of the image that can be inferred without the text information
  \item \textcolor[rgb]{0,0,0.9}{Requirement 3}: Do not use phrases like "mentioned", "caption", "context", "without the image" in the question/answer pairs. Instead, refer to the information as being "in the image" or preferably don't mention anything
  \item \textcolor[rgb]{0,0,0.9}{Requirement 4}: Ensure that question/anwer pairs are diverse and cover a range of visual aspects of the image
  \item \textcolor[rgb]{0,0,0.9}{Requirement 5}: Answer responsibly, avoiding overconfidence, and do not provide medical advice or diagnostic information
\end{itemize}

\textcolor[rgb]{0,0,0.9}{Caption:}  \{caption\}

\textcolor[rgb]{0,0,0.9}{Question:}

\textcolor[rgb]{0,0,0.9}{Answer:}

\tcblower
\textcolor[rgb]{0,0,0.9}{Model Response:} 
\scriptsize

\textcolor[rgb]{0,0,0.9}{Question:} What type of cell morphology can be observed in the image?\newline
\textcolor[rgb]{0,0,0.9}{Answer:} The image likely displays spindle-shaped cells, which are a characteristic of certain types of tumors.\\

\textcolor[rgb]{0,0,0.9}{Question:} What is the visual appearance of the cells in the image?\newline
\textcolor[rgb]{0,0,0.9}{Answer:} The cells in the image likely have a prominent eosinophilic appearance, meaning they are stained bright pink or red.\newline

\textcolor[rgb]{0,0,0.9}{Question:} What distinctive feature might some cells in the image show?\newline
\textcolor[rgb]{0,0,0.9}{Answer:} Some cells in the image might show prominent paranuclear vacuolisation, which is a clear or empty space around the nucleus of the cell.\newline

\textcolor[rgb]{0,0,0.9}{Question:} Where can we observe the fascicles of eosinophilic spindle cells in the image?\newline
\textcolor[rgb]{0,0,0.9}{Answer:} The fascicles of eosinophilic spindle cells are likely to be spread throughout the image, arranged in bundles or clusters.\newline

\textcolor[rgb]{0,0,0.9}{Question:} What similar characteristic could be inferred from the image when compared to leiomyosarcoma?\newline
\textcolor[rgb]{0,0,0.9}{Answer:} Just like in images of leiomyosarcoma, we might see prominent paranuclear vacuolisation in some of the cells in this image.
\end{tcolorbox}
\end{document}


\author{Anonymous authors}
\institute{Anonymous institutes}
\title{Supplementary Material: Path-RAG}
\maketitle  

\section{Path-RAG Prompts}
\begin{tcolorbox}[enhanced jigsaw, breakable, pad at break*=1mm, attach boxed title to top center={yshift=-4mm,yshifttext=-1mm}, 
colback=blue!5!white,colframe=blue!75!black,colbacktitle=gray!80!white,title=Prompts,fonttitle=\bfseries,boxed title style={size=small,colframe=gray!50!black}]
\small
{\textcolor[rgb]{0,0,0.9}{Generate Answer from LLaVa-Med:}} 
\begin{itemize}
    \[<question> + <image>\]
\end{itemize}
 {\textcolor[rgb]{0,0,0.9}{Get Response from GPT-4:}} You are a professional pathologist. Please generate a comprehensive answer in several sentences that integrates multiple perspectives for a pathology image, and provide a balanced conclusion based on the information provided.

\begin{itemize}
  \item \textcolor[rgb]{0,0,0.9}{Perspective 1}: \textit{Answer from LLaVa-Med with full image input}
  \item \textcolor[rgb]{0,0,0.9}{Perspective 2}: \textit{Answer from LLaVa-Med with Patch 1 input}
  \item \textcolor[rgb]{0,0,0.9}{Perspective 3}: \textit{Answer from LLaVa-Med with Patch 2 input}
  \item \textcolor[rgb]{0,0,0.9}{Perspective 4}: \textit{Answer from LLaVa-Med with Patch 3 input}
  \item \textcolor[rgb]{0,0,0.9}{Question}: \textit{Question}
\end{itemize}
  
  \tcblower
  \small
{\textcolor[rgb]{0,0,0.9}{Generate Description from LLaVa-Med:}} 
\begin{itemize}
    \["Describe~the~following~image~in~detail." + <image>\]
\end{itemize}

{\textcolor[rgb]{0,0,0.9}{Get Response from GPT-4:}} You are a professional pathologist. Please answer the following question in several sentences from the description of an image and three important patches.

\begin{itemize}
  \item \textcolor[rgb]{0,0,0.9}{Description of image}: \textit{Description from LLaVa-Med with full image input}
  \item \textcolor[rgb]{0,0,0.9}{Description of patch 1}: \textit{Description from LLaVa-Med with Patch 1 input}
  \item \textcolor[rgb]{0,0,0.9}{Description of patch 2}: \textit{Description from LLaVa-Med with Patch 2 input}
  \item \textcolor[rgb]{0,0,0.9}{Description of patch 3}: \textit{Description from LLaVa-Med with Patch 3 input}
    \item \textcolor[rgb]{0,0,0.9}{Question}: \textit{Question}
\end{itemize}

\end{tcolorbox}

\section{Path-RAG Example} 

\begin{tcolorbox}[enhanced jigsaw, breakable, pad at break*=1mm, attach boxed title to top center={yshift=-4mm,yshifttext=-1mm}, 
colback=blue!5!white,colframe=blue!75!black,colbacktitle=gray!80!white,title=Path-RAG GPT-4 Prompt and Response,fonttitle=\bfseries,boxed title style={size=small,colframe=gray!50!black}]
\small
   {\textcolor[rgb]{0,0,0.9}{Prompt Design:}} You are a professional pathologist. Please generate a comprehensive answer in several sentences that integrates multiple perspectives for a pathology image, and provide a balanced conclusion based on the information provided.

\begin{itemize}
  \item \textcolor[rgb]{0,0,0.9}{Perspective 1}: "other tubules"
  \item \textcolor[rgb]{0,0,0.9}{Perspective 2}: "other molecules"
  \item \textcolor[rgb]{0,0,0.9}{Perspective 3}: "small intestine"
  \item \textcolor[rgb]{0,0,0.9}{Perspective 4}: "red cells in vessels well shown"
\end{itemize}
  \tcblower
 {\textcolor[rgb]{0,0,0.9}{Model Response:}} 
\scriptsize
 
In the pathology image under consideration, we are presented with a complex interplay of various elements that require a detailed analysis from multiple perspectives to reach a comprehensive understanding.
 
From the first perspective, focusing on "other tubules," we observe structures that are likely to be part of the nephron system within the kidney. These tubules, when cut in cross-section, reveal a detailed architecture essential for the kidney's filtration and absorption processes. Their presence in the image suggests a renal context, which is crucial for understanding the overall pathology.

The second perspective, "other molecules," hints at the biochemical aspect of the pathology. This perspective could involve the identification of specific molecules within the tubules or surrounding tissues, which could be indicative of metabolic processes or pathological states. Molecules such as proteins, lipids, or carbohydrates, when visualized at a microscopic level, can provide insights into the cellular functions or dysfunctions present.

The third perspective, "small intestine," introduces a different anatomical context. The small intestine, characterized by its villi and microvilli, is essential for nutrient absorption. If the structures cut in cross-section are part of the small intestine, this would imply a focus on the gastrointestinal system rather than the renal system. This perspective necessitates a consideration of the enteric environment and its implications for nutrient absorption and digestion.

The fourth perspective, "red cells in vessels well shown," provides a vascular viewpoint. The presence of red cells in vessels, clearly visible in the image, underscores the importance of blood supply and circulation within the examined tissue. This aspect is crucial for understanding the tissue's health, as it relates to oxygen and nutrient delivery as well as waste removal.

Integrating these perspectives, the question of what is cut in cross-section can be approached with a balanced conclusion. Given the mention of tubules, molecules, and the small intestine, alongside the clear visualization of red cells in vessels, it is plausible that the image presents a cross-section of the small intestine, with a particular emphasis on its vascularization and microscopic structure. The tubules mentioned could refer to the intestinal \textbf{glands} (crypts of Lieberkühn) within the small intestine, which are essential for its function. The molecules could pertain to the various enzymes and hormones active in this region. The detailed visibility of red cells within the vessels highlights the rich vascular supply to the small intestine, essential for its absorptive and digestive functions. Thus, the comprehensive analysis suggests that the cross-sections likely represent the small intestine, with a detailed view of its internal structure and vascularization.

\end{tcolorbox}